\documentclass[runningheads]{llncs}
\usepackage{graphicx}
\usepackage{xcolor}
\begin{document}
\title{Advancements in synthetic data extraction for industrial injection molding}
%
%

\author{
Rottenwalter Georg\inst{1,2} \and
Tilly Marcel\inst{1,3} \and
Bielenberg Christian\inst{1,4} \and
Obermeier Katharina\inst{1, 5}
}

\authorrunning{Rottenwalter et al.}

%

\institute{Rosenheim Technical University of Applied Sciences – Hochschulstraße 1, 83024 Rosenheim
\and \email{georg.rottenwalter@th-rosenheim.de} \and \email{marcel.tilly@th-rosenheim.de} \and 
\email{christian.bielenberg@th-rosenheim.de} \and \email{katharina.obermeier@stud.th-rosenheim.de}}

\maketitle 
\begin{abstract}
Machine learning has significant potential for optimizing various industrial processes. However, data acquisition remains a major challenge as it is both time-consuming and costly. Synthetic data offers a promising solution to augment insufficient data sets and improve the robustness of machine learning models. In this paper, we investigate the feasibility of incorporating synthetic data into the training process of the injection molding process using an existing Long Short-Term Memory architecture. Our approach is to generate synthetic data by simulating production cycles and incorporating them into the training data set. Through iterative experimentation with different proportions of synthetic data, we attempt to find an optimal balance that maximizes the benefits of synthetic data while preserving the authenticity and relevance of real data. Our results suggest that the inclusion of synthetic data improves the model's ability to handle different scenarios, with potential practical industrial applications to reduce manual labor, machine use, and material waste. This approach provides a valuable alternative for situations where extensive data collection and maintenance has been impractical or costly and thus could contribute to more efficient manufacturing processes in the future.

\keywords{industry \and injection molding \and synthetic data \and machine learning \and data generation \and time series data}
\end{abstract}
\section{Introduction}
Machine learning (ML) is used in many industries for quality assurance and production optimization. In future, many production processes and other operations will be optimized by ML models. One of the most prominent challenges to optimizing processes through ML is data collection, which is one of the biggest and most critical bottlenecks. Most of the time dedicated to ML processes from start to finish is typically devoted to data preparation and procurement, encompassing tasks such as data collection, cleaning, analysis, visualization, and feature engineering \cite{roh2019survey}.
However, since large and high-quality data sets already exist in many areas of the ML world, this is not true for the industry in general. Despite the lack of data, at first glance, data acquisition in the industry should not be a problem, as countless processes can be tracked and recorded in production, which can be used to improve product quality or production optimization. Unfortunately, these processes have to be monitored, stored, and labeled, which is time-consuming and cost-intensive to be able to use the data for ML.
This possibility of data acquisition is a double-edged sword since the production cycles only produce the same products or products that are difficult to distinguish. So it is hard to grant greater variance between different states of product quality without being forced to produce broken products.
To produce many different qualities would mean a lot of deliberate production waste, many machine hours, a lot of skilled labor time, and wasted materials.
Since most companies try to keep the cost-benefit factor of their production chains as high as possible, synthetic data could be used in the future to generate or enrich larger data sets. This would provide cheap and efficient data if large datasets are not existing.
For this very purpose, we propose in our paper to enrich and augment too small or insufficient datasets with synthetically generated data. The approach is to build from existing literature by using the simulated operations of production cycles to augment our training dataset for ML. To demonstrate our approach, we have taken an existing Long Short-Term Memory (LSTM) architecture that is designed to classify the quality of an injection molding process and trained it with two data sets. 
The first data set consisted only of real data and was too small for representative training. A synthetic data set was used as the second data set. In our training, the proportion of total data was then increased with up to 30\% synthetic data. 

Our goal is to show that it is possible to enhance data from injection molding processes with a certain fraction of synthetically generated data to extend the robustness of the model. With this approach, we want to make the model more adaptable to different scenarios. In this way, we can use simulated production processes to generate production errors as well as normal processes already in the simulated environment, without having to test them on an actual production machine. 


\section{Related work}
\subsection{Synthetic data generation}

Automatic techniques for generating synthetic datasets with accompanying labels are gaining popularity in ML, owing to their cost-effectiveness and adaptability. This approach offers an efficient alternative to traditional data collection and annotation methods, streamlining the development of ML models \cite{patki2016synthetic}.
In the field of computer vision, synthetic data can already be generated very successfully via 3D modeling. For example, a purely synthetic data set of 3D models generated from real objects was trained successfully in a Convolutional Neural Network \cite{wong2019synthetic}.
Another use case for generating synthetic data is text in images. Here, fonts, sizes, and the colors of the text are varied to generate large sets of data \cite{gupta2016synthetic}.
In another method, a synthetic dataset of 2D mechanical designs, called GMCAD, was used to train a deep learning model for Design for Additive Manufacturing \cite{almasri2022gmcad}.

Similar to the data generation methods mentioned above, our approach attempts to generate synthetic data through the simulation of an injection molding process.

\subsection{Injection molding simulation}

The development of a low-cost, desktop-based virtual injection molding system that utilizes virtual reality, finite element analysis, motion simulation, and scientific visualization techniques sought to reduce the number of costly iterations of physical test molding by determining mold structure and process parameters. This system was intended to provide engineers with a comprehensive view of mold structure and assembly while identifying potential defects during the molding process, ultimately improving moldability and product quality \cite{zhou2009virtual}.
In another study, a 3D simulation of mold filling for high-density polyethylene in injection molding was performed using ANSYS-CFX, as well as injection molding experiments with deliberately short shots at different optimal metering percentages for melt front tracking. The comparison between the simulation and experimental results shows reasonable agreement in terms of the shape of the melt front, with minor variations due to the nature of the injection molding experiments \cite{mukras2016simulation}.
It is necessary that the inputs and outputs of an injection molding simulation match as closely as possible the actual conditions of a real injection molding machine. By benchmarking injection molding simulations, the importance of matching simulation inputs with actual conditions of real injection molding machines will be discussed. The study examines factors that affect simulation accuracy, such as part geometry, model mesh type, material data, and process settings, as well as validation techniques using transducers and part deflection measurement methods \cite{speight2008best}.

The preceding approaches are intended to give a small indication of the methods or approaches available in the field of injection molding simulation. Our goal will be to proceed similarly to the studies mentioned above to have our processes recorded by a simulation.

\subsection{ML in injection molding}

The integration of ML techniques in injection molding processes presents opportunities for enhancing optimization and achieving superior performance.
For example, Artificial Neural Network (ANN)-based temperature controller design was developed for a plastic injection molding system. For this purpose, an integrator and an ANN were combined. The proposed controller demonstrates improved performance in terms of reduced overshoot and faster transient response compared to an industrial proportional integral derivative regulator \cite{khomenko2019ann}.
In another project, an ANN-based position controller for an all-electric injection molding machine was presented to improve dynamic positioning characteristics for various components. The ANN was trained with experimental data and backpropagation algorithms to provide accurate control of motor current, position, and velocity, and its efficiency was verified using real-world time series data \cite{veligorskyi2019artificial}.

Another important approach in the ML injection molding environment is the approach to improve the accuracy of fault detection models in manufacturing processes. There, error rates are typically very low, leading to unbalanced data sets. By using ML and sharing process data between multiple injection molding work systems, model performance can be improved using industry data from approximately two million process cycles \cite{kozjek2017knowledge}.

Using similar ML approaches, we are trying to train a neural network to recognize qualities in the injection molding process. Since small and unbalanced data sets are not sufficient to train representative models, we will try to scale up our training data using synthetic data.

\section{Methods}

\begin{figure}
\includegraphics[width=\textwidth]{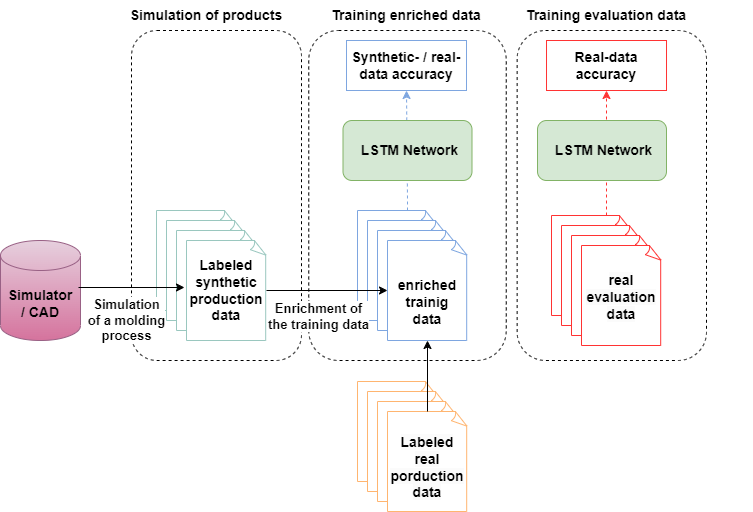}
\caption{System design of overall concept.} \label{fig1}
\end{figure}

Figure \ref{fig1} shows the system structure of our concept of data generation up to training. The first point \textit{Simulation of products} is responsible for the generation of the synthetic data. In the step \textit{Training enriched data}, the synthetic part of the training sets is determined and trained with a certain part of synthetic data. In the last step, \textit{Training evaluation data} is evaluated and used to compare the model with real data.

\begin{itemize}
  \item \textbf{Simulation of production processes:} This step mainly consists of simulating a real production process on the injection molding machine and storing it in a time series-based file. The data set should store all the values and parameters from the start of the cycle to the end of the cycle.
  \item \textbf{Labeling the simulated product cycles:} In this stage, all collected production cycles are classified and labeled by an already existing classifier. So after this step, each cycle file should have a label with e.g. the production quality.
  \item \textbf{Enrichment of training sets:} To get a reasonable percentage of synthetic data in the real training data, different percentages were tried.
  \item \textbf{Testing and evaluation:} In order to compare the data sets, an LSTM network was trained with the different data, and the accuracy was compared on an independent evaluation dataset.
\end{itemize}

\subsection{Simulation of production processes}

In the context of generating synthetic data for injection molding processes, a similar approach can be taken as in the previously mentioned research areas such as 3D modeling and text in images. To generate synthetic data sets in a simulated injection molding environment, one can start by collecting real data from actual injection molding processes, which may include process parameters, machine settings, and part geometries.

In our case, a CAD program and a production simulator serve as a virtual environment to generate synthetic data by changing parameters and introducing variability in process settings, part geometries, and material properties. In this way, a simulation model can be developed based on this real data to replicate the injection molding process in CAD and take into account key aspects such as material flow, cooling, and pressure distribution. In order to include all other boundary parameters of this production process, a production simulator, normally used for training purposes, was used to simulate the entire process. 
The simulator receives setting parameters, such as injection speed, changeover point, holding pressure level, and all common parameters of an injection molding machine. This results in the process parameters such as screw position, residual compound cushion, ejector speed, and many more. The simulator is a clone of the machine control and thus offers the possibility to use all setting parameters of a real machine and generate the production data based on them. All production data are output, but the values are not close to reality, which is why the CAD program mentioned above is used.
In order to obtain a complete process as a time series file, the actual process of injecting the plastic was simulated in CAD, and in addition, the operating part and machine process were run on a production simulator.

\subsection{Labeling the simulated product cycles}

In connection with the generation of synthetic data for injection molding processes, the CAD-based simulation model was extended by integrating a production simulator for generating time series data. This simulator allows us to generate synthetic data that capture the temporal dynamics of the injection molding process and provides valuable information for data-driven modeling and analysis. To obtain accurate labels for the generated time series data, we use our expertise and knowledge of the parameters set in the production simulator, as well as the output quality of the CAD fill study. This manual labeling approach ensures that the synthetic data is appropriately labeled, allowing effective use in ML applications. Alternatively, we could use an already trained simulator to automatically label the data, streamlining the process and increasing the overall efficiency of the data generation pipeline. An important aspect of our synthetic data generation methodology is to intentionally generate a larger percentage of non-compliant or defective products within the simulation. This is done to compensate for the inherent imbalance in real production data, where such cases are generally rare. By generating more diverse and balanced datasets that contain a larger number of non-conforming products, we can improve the performance of ML models, especially in tasks related to defect detection and process optimization.

\subsection{Enrichment of training sets}

To effectively incorporate synthetic data into the training process and improve the performance of ML models, it is critical to determine an appropriate balance between synthetic and real data in the training sets. The goal is to strike a balance between taking advantage of synthetic data and maintaining the authenticity and relevance of real data.
Through iterative experimentation with different proportions of synthetic data, the goal is to find an appropriate balance that maximizes the benefits of synthetic data, such as improved model robustness and generalizability, while preserving the essential features of the real production data.

\subsection{Testing and evaluation}

In this section, we investigate the impact of different proportions of synthetic data on the overall performance of ML models, specifically using an LSTM network. LSTM networks are an advanced alternative to traditional recurrent neural networks. They have become the first choice for processing time-series data because of their ability to effectively manage long-term dependencies in sequential data such as text, audio, and video \cite{yu2019review}.
We systematically evaluate different ratios of synthetic to real data in training sets, ranging from minimal inclusion of synthetic data to a significant amount. During this process, the LSTM models are trained and validated with training sets consisting of different mixtures of synthetic and real data. For a comparative analysis, we evaluate the accuracy of LSTM models trained on semi-synthetic datasets compared to a validation set containing only real data. This approach allows us to measure the effectiveness of including synthetic data in the training process and determine an optimal ratio that maximizes model performance. Insights gained from this analysis help fine-tune the composition of training sets and ultimately contribute to improving the performance of ML models, particularly in tasks related to fault detection and process optimization.

\section{Experimental setup}

\subsection{Real training \& validation data}

For our study, a data set of 275 injection molding processes was used as training data. An experienced injection molder evaluated the quality of the parts and recorded the relevant information for each production cycle. To ensure a wide range of processes, we intentionally varied key process parameters such as piston stroke, rounds per minute in odization, injection volume, back pressure, holding pressure, and other relevant factors. Given the inherent time intervals between sensor samples in the data, we chose a data augmentation strategy to improve the training data. The data set had a consistent interval of 10 ms between successive sensor samples within the time series. We enlarged each dataset by quadrupling it and stored each first through the fourth step at 10 ms intervals in separate files. Thus, the expanded data set included a total of 1100 labeled production cycle data points. These are divided into two classes with 56.5\% good cycles and 43.5\% not good cycles.

As data for the validation process for the trained model, 33\% of the real data is taken. This corresponds to 363 data sets, which are shuffled in each training. This evaluation data set is important because it is separate and allows us to independently verify the results from the training and compare them to the data. Thus, the real data is split between 737 training data and 363 validation data.

\subsection{Synthetic data}

\begin{figure}
\includegraphics[width=\textwidth]{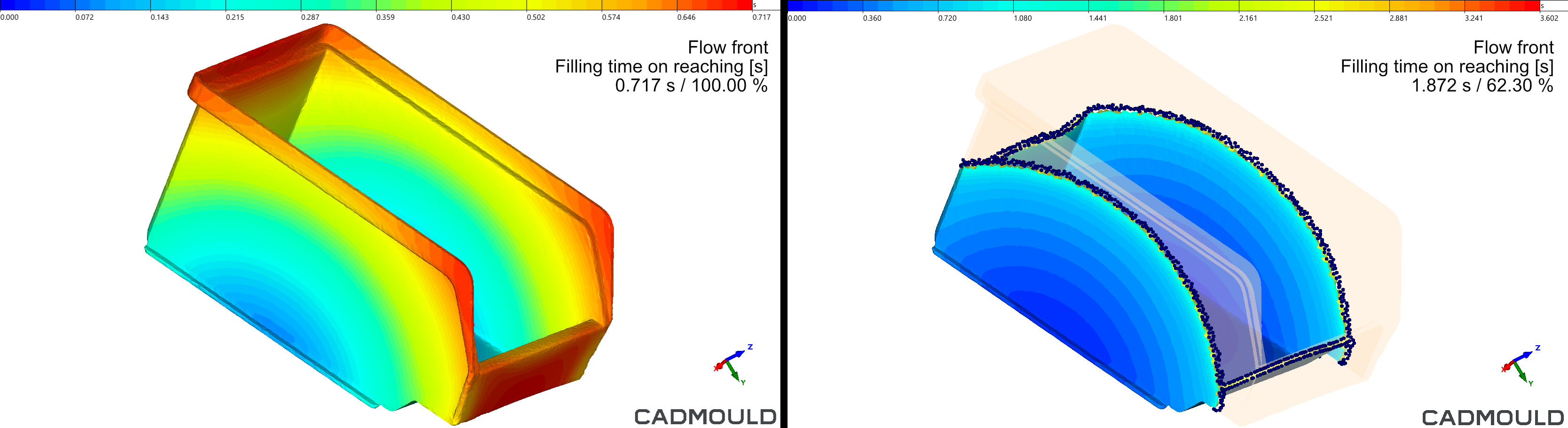}
\caption{A pictorial example of the evaluation of a CAD simulated filling study. \textit{The production piece on the left is good, the piece on the right is rated as not good because it is not sufficiently filled.}} \label{fig2}
\end{figure}

For the synthetic dataset, 100 synthetic injection molding cycles were generated. These are divided into two classes 60\%, not good cycles, and 40\% good cycles. The 100 data sets were also quadrupled like the real data to augment them to 400 data sets. As explained in the Methods, the synthetic data were evaluated using the CAD output to assign the correct part quality corresponding to the class in the synthetic data set. Figure \ref{fig2} illustrates the part qualities that were evaluated using the CAD program. 
Furthermore, in addition to the illustration, critical quality indicators and optimal parameters in the injection molding process such as material flow, cooling dynamics, and pressure distribution are displayed in the CAD program.

\subsection{LSTM training}

The LSTM for our neural network, which is designed to process time series data, consists of 34 input neurons that comprise our feature in the injection molding process, one output neuron, and three LSTM layers. After each LSTM layer, a dropout layer is added to avoid overfitting. The first two LSTM layers have the return\_sequences attribute set to True so they can maintain their state over multiple time steps.
The paper "Sequence to Sequence Learning with Neural Networks" served as the approach for our network \cite{sutskever2014sequence}. Here, the effectiveness of deep LSTM networks in handling complex tasks was demonstrated, which led us to use a similar architecture.
Our LSTM network is designed for classification and follows a unidirectional many-to-one configuration \cite{smagulova2019survey}. This configuration allows the network to process a sequence of input data and produce a single output suitable for our classification task.
Table~\ref{tab1} provides a detailed overview of the specific settings and variables used in our LSTM network, including layer configurations, activation functions, and other hyperparameters. By carefully selecting and tuning these settings, we aim to create an LSTM network that can provide accurate classification results for our time series data.
It's important to note that we did not conduct an extensive comparison with other LSTM architectures for this study. The selected model was the best performer from our initial round of experimentation and seemed promising for our specific task.

The system used for all computation had a 12th Gen Intel Core i9-12900KS 3.40 GHz CPU with 64GB memory and an NVIDIA GeForce RTX 3090 Ti GPU. As an environment, Tensorflow 2.11.0 and Keras 2.11.0 were used. 

\begin{table}
\centering
\caption{Variables for network training that have remained constant}\label{tab1}
\begin{tabular}{|l|l|l|}
\hline
Learning rate & 0.0001175\\
Input dimension & 34\\
Output size & 1\\
Units LSTM layer 1 & 100\\
Units LSTM layer 2 & 100\\
Units LSTM layer 3 & 100\\
Dropout rate behind LSTM layer 1 and 2 & 0.1598\\
Dropout rate after last LSTM layer & 0.279\\
Optimizer & Adam\\
Loss & MSE\\
Batch size & 64\\
Epochs & 50\\
\hline
\end{tabular}
\end{table}

\section{Results}

\subsection{Quality of classification}

In the following section, we present the results of our experiments on the classification of component qualities with different proportions of synthetic data. We conducted 7 training sessions with 50 runs each, where the proportion of synthetic data in the training data set ranged from 0\% to 30\% and was increased in increments of 5\%.

In a subsequent training iteration, the synthetic data was increased in 5\% increments within the overall data set. However, the overall size of the data set was kept constant to analyze the effects of a lower ratio of real data to synthetic data.

The goal of these experiments was to analyze the relationship between the inclusion of synthetic data in the training dataset and the performance of our LSTM model in terms of loss and accuracy. For each training session, the maximum and minimum values of loss and accuracy were evaluated and the average values were calculated for each of the 50 runs.

For example, for the test runs without synthetic data, the average validation accuracy was 93,6\%, while the training accuracy was 92.8\%. When 30\% of synthetic data was added to the dataset in the first experimental run, the average validation accuracy dropped slightly to 92.3\% and the training accuracy to 85.5\%.

A detailed overview of the experimental results, including the different accuracy values and the distribution of synthetic data in the training, can be found in Table \ref{tab2}. Since the percentages are based on the total data set, Table \ref{tab2} was expanded to include the column \textit{"synthetic fraction training set"} because the training test split reduced the size of the real training set, and thus the percentage of synthetic data in the training had to be adjusted.

\begin{table}
\centering
\caption{The distribution of the synthetic data added to the training set with the mean results from the validation data, training data, F1 score and AUC-ROC }\label{tab2}
\begin{tabular}{|l|l|l|l|l|l|l|l|l|l|}
\hline
\multicolumn{4}{|c|}{added synthetic data} &                                        
\multicolumn{6}{c|}{mean values} \\ 
\hline
\multicolumn{2}{|c|}{total set} &                                        
\multicolumn{2}{|c|}{training set} &                                         
\multicolumn{2}{|c|}{validation} &                                        
\multicolumn{2}{|c|}{training} &                                        
\multicolumn{2}{|c|}{performance meas.}  \\ 
\hline
\% & count & \%  & count & accuracy & loss & accuracy & loss & F1 score & AUC-ROC \\
\hline
0 \% &  1100 &  0 \%     & 737 & 0.9363 & 0.05935 & 0.9284 & 0.06972 & 0.9238 & 0.9743 \\
5 \% &  1155 &  7.5 \%   & 792 & 0.9164 & 0.07419 & 0.8918 & 0.09062 & 0.8979 & 0.9637 \\
10 \% & 1210 &   15 \%   & 847 & 0.9221 & 0.07223 & 0.8795 & 0.09691 & 0.907 & 0.9673   \\
15 \% & 1265 &   22.4 \% & 902 & 0.9007 & 0.087 & 0.8464 & 0.1168 & 0.8784 & 0.9542   \\
20 \% & 1320 &   29.9 \% & 957 & 0.9165 & 0.07609 & 0.8548 & 0.1089 & 0.9009 & 0.9633   \\
25 \% & 1375 &   37.3 \% & 1012 & 0.9139 & 0.07549 & 0.8515 & 0.11 & 0.8945 & 0.963   \\
30 \% & 1430 &   44.8 \% & 1067 & 0.9233 & 0.07094 & 0.8523 & 0.1106 & 0.908 & 0.9677   \\

\hline
\end{tabular}
\end{table}

In the subsequent training phase, the overall size of the data set was kept constant while the proportion of synthetic data was adjusted. The goal of these experiments was to investigate the effects of adding synthetic variants to a smaller subset of real data. We started with a small subset of the real data and investigated whether the integration of synthetic data could improve performance results.
Therefore, we first started the training with a smaller amount of synthetic data and then gradually filled it up with synthetic data until it was equal to the volume of our original real data set used as a benchmark for training.
When the training set consisted of 55.2\% real data derived from the original dataset and contained no synthetic data, we achieved an average validation accuracy of 82.7\% and an average training accuracy of 82\%. In contrast, when 44.8\% of synthetic data was added to the dataset, the average validation accuracy increased to 83.9\%, but the average training accuracy decreased to 75.5\%. Table \ref{tab3} shows these different metrics and represents a marginal deviation from our primary experimental framework.

Analyzing these results, we found that the training accuracy decreased as the proportion of synthetic data increased. The validation accuracy remained relatively stable despite the inclusion of synthetic data. This observation suggests that the inclusion of additional synthetic data in the training process does not significantly affect the overall accuracy of the model, but contributes to its robustness to potential perturbations.

\begin{table}
\centering
\caption{The distribution of the real and synthetic data in the training set with the mean results from the validation data, training data, F1 score and AUC-ROC }\label{tab3}
\begin{tabular}{|l|l|l|l|l|l|l|l|l|l|}
\hline
\multicolumn{4}{|c|}{training data} &                                        
\multicolumn{6}{c|}{mean values} \\ 
\hline
\multicolumn{2}{|c|}{real data} &                                        
\multicolumn{2}{|c|}{synthetic data} &                         
\multicolumn{3}{|c|}{real data} &                                        
\multicolumn{3}{|c|}{synthetic data}  \\ 
\hline
\% & count & \%  & count & val. acc & train. acc & F1 score & val. acc & train. acc & F1 score \\
\hline
100 \%  &  737 &  0 \%        & 0   & 0.9363 & 0.9284 & 0.9238 & - & - & - \\
92.5 \% &  682 &  7.5 \%   & 55  & 0.919 & 0.9143 & 0.9032  & 0.9144 & 0.8904 & 0.8965 \\
85 \%   & 627 &   15 \%      & 110 & 0.906 & 0.904 & 0.8855   & 0.9123 & 0.8666 & 0.8941   \\
77.6 \% & 572 &   22.4 \%  & 165 & 0.8889 & 0.882 & 0.8636  & 0.9059 & 0.8485  & 0.887  \\
70.1 \% & 517 &   29.9 \%  & 220 & 0.8173 & 0.8201 & 0.7636 & 0.8755 & 0.8076  & 0.8475  \\
62.7 \% & 462 &   37.3 \%  & 275 & 0.8491 & 0.8509 & 0.8034 & 0.8576 & 0.783  & 0.8182  \\
55.2 \% & 407 &   44.8 \% & 330 & 0.8273 & 0.82 & 0.7745    & 0.8397 & 0.755  & 0.7922  \\

\hline
\end{tabular}
\end{table}

\begin{figure}
\includegraphics[width=\textwidth]{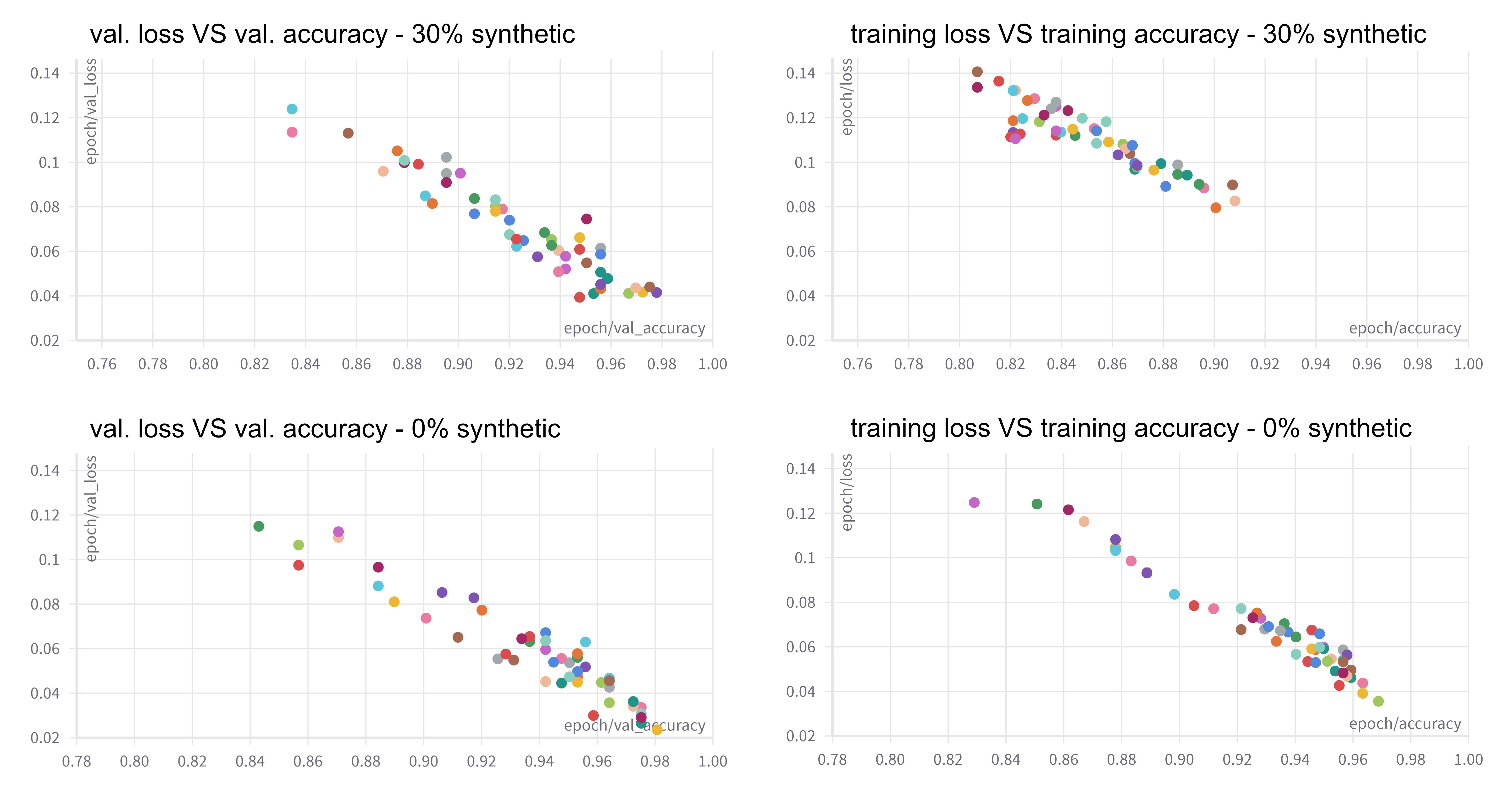}
\caption{While the validation accuracies and losses with 30\% synthetic data \textit{(top left)} and 0\% synthetic data \textit{(bottom left)} remain relatively the same, one can see that the training accuracy loss graph scatters much more for a dataset with 30\% synthetic data \textit{(top right)} than to a dataset with 0\% \textit{(bottom right)}.} \label{fig3}
\end{figure}

Figure \ref{fig3} illustrates the robustness of the model from the first experimental run to perturbations by comparing the validation accuracies and losses for data sets with 0\% and 30\% synthetic data. It can be observed that the validation accuracy loss graph is relatively similar in both cases, but the training accuracy loss graph shows a larger variation for the dataset with 30\% synthetic data than for the dataset with 0\% synthetic data.  Moreover, the training accuracies and losses decrease more, as shown by the shifted values in the graph.

The robustness of our model from the first experimental run is remarkable given the inherent unpredictability of synthetic data. Despite these challenges, our model showed an small change in the loss and accuracy metrics, as mentioned earlier. This minimal impact was reflected in only negligible changes in the F1 score and the area under the receiver operating characteristic curve (AUC-ROC). Thus, Table \ref{tab2} shows that the F1 score is 0.9238 for 0\% synthetic data and only 0.908 for 30\% synthetic data as another example, the ROC-AUC is listed, which is 0.9743 for 0\% synthetic data and 0.9677 for 30\% synthetic data. This proves that our model maintained not only its robustness but also its predictive capabilities, confirming the usability of synthetic data in data analysis and machine learning tasks.

Table 3 demonstrates that the F1 score is improved with the inclusion of more than 15\% synthetic data compared to an absence of synthetic data in our dataset. The results indicate that with a smaller training set, we can slightly enhance validation accuracy, even though the training accuracy is somewhat inferior compared to when the training solely relies on real data.
In our second experimental iteration, we illustrate that the high quality of our data enables us to not only strengthen the model with minimal training data but also effectively augment it with data that marginally improves the training, even if just by a single percentage point. Therefore, we observe that the validation accuracy of a modest dataset, when supplemented with synthetic data, surpasses that of a non-augmented dataset.

\section{Evaluation}
Our study provided us with valuable insights into the use of synthetic data, but we still had two major limitations in our approach.

First, the amount of real-world data available for training was insufficient to create a fully representative dataset. This limitation resulted in the LSTM network achieving an average validation accuracy of about 94\% on the real data set, which is a satisfactory result for the beginning, but cannot be generalized due to the small amount of data.
Another limitation related to the process of generating synthetic data using CAD simulation. The labeling of the injection molding cycles had to be done manually, as our automated method for this crucial step had not yet been worked out. This manual approach increased the overall time and effort required for data preparation.

Future research should attempt to address these limitations by incorporating larger, more representative real-world datasets and automating the labeling process to generate synthetic data.

\section{Conclusion}

We have shown that it is possible to train injection molding processes with a combination of real and synthetic data to improve the robustness of the resulting models. Our results suggest that the inclusion of synthetic data in the training process not only improves the model's ability to handle different scenarios but also offers promising potential for practical industrial applications.
By incorporating synthetic data, we can improve the performance and generalization ability of ML models.

Thus, in the next studies, we will try to increase the synthetic fraction in our data as much as possible. In addition, we will try to use Generative Adversarial Networks (GANs) to generate more synthetic data that also take into account the noise and uncertainties of real-world data 
  \cite{goodfellow2020generative} \cite{goodfellow2016nips}.
For a less complex label process to evaluate the production cycles, we will try to train an independent classifier in the future that can label the synthetic data. This will allow us to spend more time augmenting the data and start trying to train a GAN that will generate data without simulation ,which holds the potential to generate more realistic, high-quality synthetic data.


By applying our approach, it may be possible to reduce the need for manual labor, machine usage, and material waste associated with traditional injection molding process optimization. In addition, our method offers a valuable alternative in situations where extensive data collection and maintenance has been impractical or costly. This can be beneficial to any company that does not have the resources to invest in extensive data collection.

We hope our approaches will pave the way for more sustainable and efficient manufacturing processes in the future.

%
%
%
%
\bibliographystyle{splncs04}
\bibliography{bibliography}
\end{document}